\documentclass[10pt,twocolumn, letterpaper]{article}
\usepackage{iccv}
\usepackage{times}
\usepackage{epsfig}
\usepackage[dvipsnames]{xcolor}
\usepackage{colortbl}
\usepackage{graphicx}
\usepackage{amsmath}
\usepackage{amssymb}
\usepackage{booktabs}
\usepackage{color}
\usepackage{multirow}

\usepackage{arydshln}

\usepackage{adjustbox}
\usepackage{bm}
\usepackage{nicefrac}
\definecolor{gray}{RGB}{238, 238, 238}

\usepackage[pagebackref=true,breaklinks=true,colorlinks,bookmarks=false]{hyperref}

\iccvfinalcopy 

\ificcvfinal\fi
\begin{document}

\title{Video Instance Segmentation in an Open-World}

\author{Omkar Thawakar\textsuperscript{1} \hspace{.1cm} Sanath Narayan\textsuperscript{2} \hspace{.1cm} Hisham Cholakkal\textsuperscript{1} \hspace{.1cm} Rao Muhammad Anwer\textsuperscript{1,3}\\ Salman Khan\textsuperscript{1} Jorma Laaksonen\textsuperscript{3} Mubarak Shah\textsuperscript{4} Fahad Shahbaz Khan\textsuperscript{1,5} \hspace{.1cm} 
\\
\textsuperscript{1}Mohamed bin Zayed University of AI, UAE \hspace{.1cm} \textsuperscript{2}Technology Innovation Institute, UAE
\hspace{.1cm}  \\ \textsuperscript{3}Aalto University, Finland  \hspace{.1cm} \textsuperscript{4}University of Central Florida, USA \hspace{.1cm} \textsuperscript{5}Link{\"o}ping University, Sweden\\
}
\maketitle
\ificcvfinal\thispagestyle{empty}\fi

\begin{abstract}
   Existing video instance segmentation (VIS) approaches generally follow a closed-world assumption, where only seen category instances are identified and spatio-temporally segmented at inference. Open-world formulation relaxes the close-world static-learning assumption as follows: (a) first, it distinguishes a set of known categories as well as labels an unknown object as `unknown' and then (b) it incrementally learns the  class of an unknown as and when the corresponding semantic labels become available. We propose the first open-world VIS approach, named OW-VISFormer, that introduces a novel feature enrichment mechanism and a spatio-temporal objectness (STO) module. The feature enrichment mechanism based on a light-weight auxiliary network aims at accurate pixel-level (unknown) object delineation from the background as well as distinguishing category-specific known semantic classes. The STO module strives to generate instance-level pseudo-labels by enhancing the foreground activations through a contrastive loss. Moreover, we also introduce an extensive experimental protocol to measure the characteristics of OW-VIS. Our OW-VISFormer performs favorably against a solid baseline in OW-VIS setting. Further, we evaluate our contributions in the standard fully-supervised VIS setting by integrating them into the recent SeqFormer, achieving an absolute gain of 1.6\% AP on Youtube-VIS 2019 val. set. Lastly, we show the generalizability of our contributions for the open-world detection (OWOD) setting, outperforming the best existing OWOD method in the literature. Code, models along with OW-VIS splits are available at \url{https://github.com/OmkarThawakar/OWVISFormer}.

\end{abstract}

\section{Introduction}
Video instance segmentation (VIS) strives to simultaneously classify, segment and track all object instances from a set of semantic classes in a given video. The problem is challenging since a diverse set of objects are desired to be accurately tracked and segmented despite real-world issues such as, fast motion, large intra-class variation and background clutter. Most existing VIS approaches~\cite{seqformer,vistr,CROSS-VIS,PCAN} typically follow a close-world assumption, \ie, all object categories to be detected are provided during the training and only seen object classes are spatio-temporally segmented at inference. \Eg, existing VIS methods evaluated on the popular Youtube-VIS benchmark~\cite{YouTube-VIS-2019,YouTube-VIS-2021} assume that annotated (known) instances of all 40 semantic categories to be segmented and tracked are available during training. Here, such a training scheme treats unannotated (unknown) objects as background. Therefore, the closed-world assumption poses issues to existing VIS methods when recognizing novel (unknown) object class instances.

The open-world problem formulation~\cite{joseph2021towards,ow-detr} relaxes the closed-world assumption by enabling the VIS model at each training episode to identify unknown object category instances as belonging to the `unknown' class while simultaneously learning to spatio-temporally segment a given set of `known' objects. Afterwards, these identified unknowns can be passed to an oracle, which annotates a set of object categories of interest. Then, the VIS model takes into account these new knowns to incrementally update its knowledge without requiring to be retrained from scratch using the previous known object categories. However, such an open-world formulation poses additional issues over the standard VIS problem challenges by requiring the model to also (i) distinguish unknown objects \textit{and} (ii) recognize them later with the arrival of progressive training data, in a unified manner. Although the open-world setting has been explored recently for detection~\cite{joseph2021towards,ow-detr} and image segmentation~\cite{learn2det_ow,wang2022open}, to the best of our knowledge, we are the first to investigate the problem and introduce a novel approach for \textit{open-world video instance segmentation} (OW-VIS). 

\begin{figure*}[t!]
    \centering
      \includegraphics[width=1\linewidth]{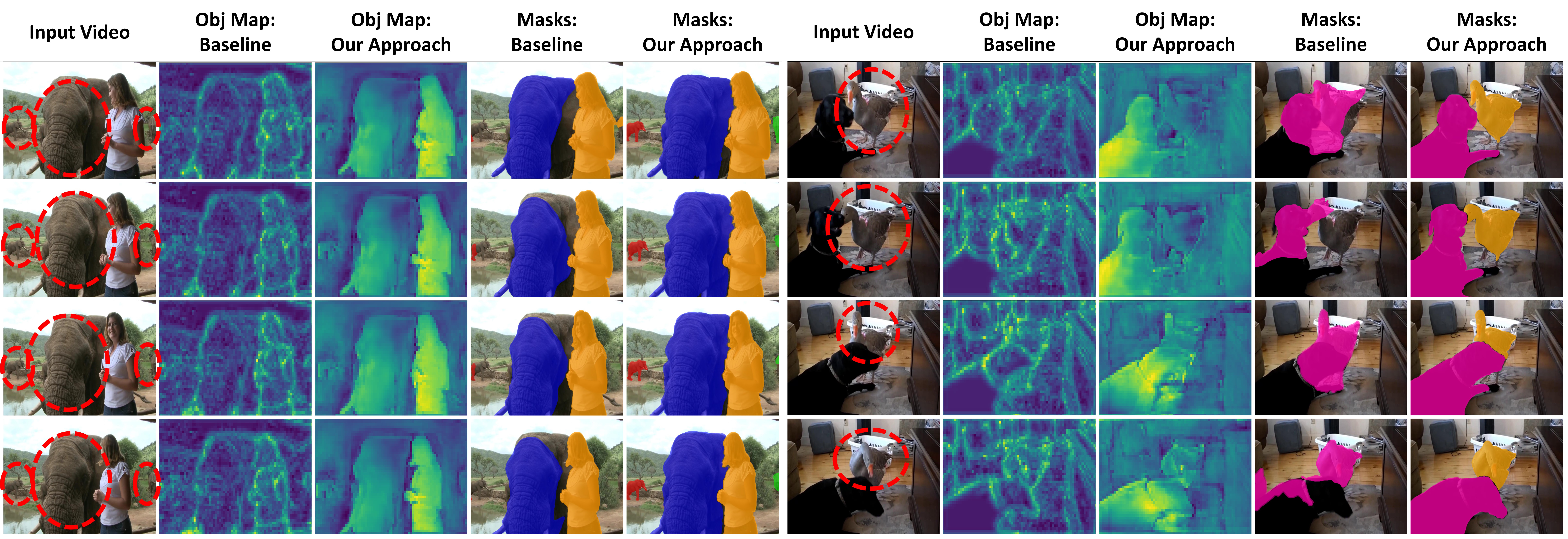}
    \caption{\textbf{Video instance segmentation in an open-world (OW-VIS) setting illustrating the first step of identifying `known' and `unknown' objects.} For each example video frame, we show the corresponding objectness map obtained from backbone features in the case of the baseline (col 2 and 7) and the spatio-temporal objectness (STO) module output map for our OW-VISFormer (col 3 and 8). 
    Moreover, we show the respective output segmentation mask for each frame in case of the baseline (col 4 and 9) and OW-VISFormer (col 5 and 10). In these example videos, the unknown (in red dashed line) objects are three \textit{elephants} (on the left), and the \textit{duck} (on the right). The known objects in these videos are \textit{person} (on the left) and \textit{dog} (on the right). Compared to the baseline, OW-VISFormer accurately segments all the `unknown' and `known' object instances. Best viewed zoomed in. Additional results are presented in the supplementary. 
    \vspace{-0.3cm}
    } 
    \label{fig:intro}
\end{figure*}

When designing an OW-VIS framework, one plausible way is to extend a fully-supervised VIS approach by introducing a pseudo-labeling scheme to identify potential unknown objects. These potential pseudo-unknowns along with the ground-truth known instances can then be utilized to learn a foreground-background class-agnostic separation as well as performing class-specific known \vs unknown instance classification. Existing fully-supervised VIS approaches typically employ an ImageNet~\cite{ImageNetVID} pre-trained classification backbone for multi-scale feature extraction to be used in the encoder. The same features can also be utilized in a bottom-up pseudo-labeling scheme in OW-VIS. However, such a pre-trained classification-based framework is likely to struggle in the OW-VIS paradigm (see Fig.~\ref{fig:intro}), where the aim is to accurately distinguish a class-agnostic unknown object from the background as well as class-specific known categories at the \textit{pixel-level}. To achieve such an accurate pixel-level (unknown) object delineation from the background, we argue that dedicated shallow features are especially desired to complement the high-level semantic pre-trained features. Moreover, since the selection of pseudo-unknowns relies on the activations in the selected feature map, it is further desired to enhance their strengths in the foreground regions (known and unknown) for learning better objectness priors.

\subsection{Contributions}
We propose an OW-VIS approach, named OW-VISFormer, that introduces (\textit{i}) a feature enrichment mechanism, which aims to better differentiate class-agnostic foreground \vs background as well as aid in  class-specific known \vs unknown instance classification and (\textit{ii}) a spatio-temporal objectness (STO) module that strives to identify candidate pseudo-unknowns. The feature enrichment mechanism is based on a light-weight auxiliary network that is trained from \textit{scratch} and generates dedicated shallow features to complement the high-level semantic standard \textit{pre-trained} features. The resulting extended features are enriched by the encoder and then use in the STO module. Our STO module employs a contrastive loss that distinguishes candidate pseudo-unknowns from the background by enhancing the foreground activations. As a result, improved video instance mask predictions are obtained for both the known and unknown classes (see Fig.~\ref{fig:intro}). Furthermore, we introduce carefully curated open-world splits of Youtube-VIS dataset for a rigorous evaluation of OW-VIS problem.

Our extensive quantitative and qualitative evaluations demonstrate the effectiveness of the proposed OW-VISFormer leading to consistent improvement in performance, compared to the baseline. In addition, we also validate our proposed contributions in the standard fully-supervised VIS problem setting by introducing them into the recent SeqFormer~\cite{seqformer}, achieving an absolute gain of 1.6\% in overall AP on the Youtube-VIS 2019 val. set. Lastly, we demonstrate the  generalizability  of our two contributions for the open-world detection (OWOD) problem setting by integrating them into the recent OW-DETR~\cite{ow-detr}. On the challenging MS COCO OWOD split, our approach outperforms the recent OW-DETR on all the tasks for both the `known' and `unknown' categories. 
\section{Open-world Video Instance Segmentation}

\subsection{Problem Formulation\label{sec:problem_formulation}} 
Let  $\mathcal{D}^t = \{\mathcal{V}^t, \mathcal{Y}^t\}$ be a progressive dataset at time $t$ containing $N_t$ videos $\mathcal{V}^t = \{V_1, \cdots, V_{N_t}\}$ with corresponding labels $\mathcal{Y}^t = \{\bm{Y}_1,\cdots,\bm{Y}_{N_t}\}$. Here, $V_i \in \mathcal{R}^{L_i \times 3 \times H \times W}$ denotes a video of length $L_i$ frames with spatial resolution $H\times W$, while  $\bm{Y}_i = \{\bm{y}_1,\cdots,\bm{y}_K\}$ denotes the ground-truth mask  annotations  of a set of $K$ object instances present in the video.  Here, $\bm{y}_j \in \mathcal{R}^{L_i \times h \times w}$ denotes the set of masks predicted for an object instance $j$ in $L_i$ frames of a video $V_i$. Let  $\mathcal{K}^t=\{1,2,\cdots,C\}$ denote the known object categories at time $t$ and $\,$ $\mathcal{U}^t =\{C\!+\!1,\cdots\}$ be a set of unknown classes that are likely to be encountered at test time. 

As discussed earlier, our OW-VIS first learns a model $\mathcal{M}^t$ that can spatio-temporally segment an unseen class instance at time $t$ as belonging to the unknown class (denoted by label $0$) in addition to segmenting the instances of previously encountered  known classes $\mathcal{K}^t$. Next, a set of these unknown instances $\bm{U}^t$ identified by $\mathcal{M}^t$ are then taken as input to  an oracle, which labels $n$ novel classes of interest and provides  new training examples for the corresponding $n$ classes. Then,  these  $n$ classes are considered as known  and  added  to the previously known $C$ classes, such that $\mathcal{K}^{t+1} = \mathcal{K}^t + \{C+1,\cdots,C+n\}$. Then, $\mathcal{M}^t$ is incrementally trained to obtain an updated model $\mathcal{M}^{t+1}$, which can spatio-temporally segment all object instances belonging to classes in $\mathcal{K}^{t+1}$ without forgetting the previously learned classes in $\mathcal{K}^{t}$. This cycle of spatio-temporally segmenting unknown instances and incremental learning of new knowledge continues over the model's life-time. 

\subsection{Baseline OW-VIS Framework}
\label{sec:baseline}

We base our approach on the recent fully-supervised (FS) SeqFormer~\cite{seqformer}. It utilizes a standard pre-trained backbone network for multi-scale feature extraction followed by a deformable transformer~\cite{Zhu_DeformableDETR_ICLR_2021} and a segmentation block for video instance mask prediction. Here, an $M$-frame video clip $\mathbf{v} \subset V_i$ is input to a pre-trained backbone network. Then, the resulting multi-scale features of each frame are input to an encoder that outputs feature maps of the same size as the input. The encoder output features together with $q$ learnable instance query embeddings $\bm{Q}^I$ are input to the decoder. Consequently, the decoder outputs $q$ instance features $\bm{F}^I$  that are then used for video mask prediction. 

A straightforward way to extend the above FS SeqFormer to OW-VIS (Sec.~\ref{sec:problem_formulation}) is to introduce a pseudo-labeling scheme for selecting potential unknown objects followed by learning to categorize these identified pseudo-unknowns into a single unknown class. One way to design such a pseudo-labeling scheme is to utilize object proposals\footnote{Proposals are obtained from the instance features $\bm{Q}^I$ and only those remaining after selecting the ground-truth class instances through Hungarian matching are considered.} having high activations in their corresponding regions of the backbone feature maps as candidates for unknown class, as in \cite{ow-detr}. These pseudo-unknowns can be then used along with ground-truth known instances to learn a foreground-background \textit{class-agnostic} separation as well as perform \textit{class-specific} known \vs unknown instance classification. We refer it as our baseline OW-VIS framework. 

\begin{figure*}[t!]
    \centering
      \includegraphics[width=0.99\linewidth]{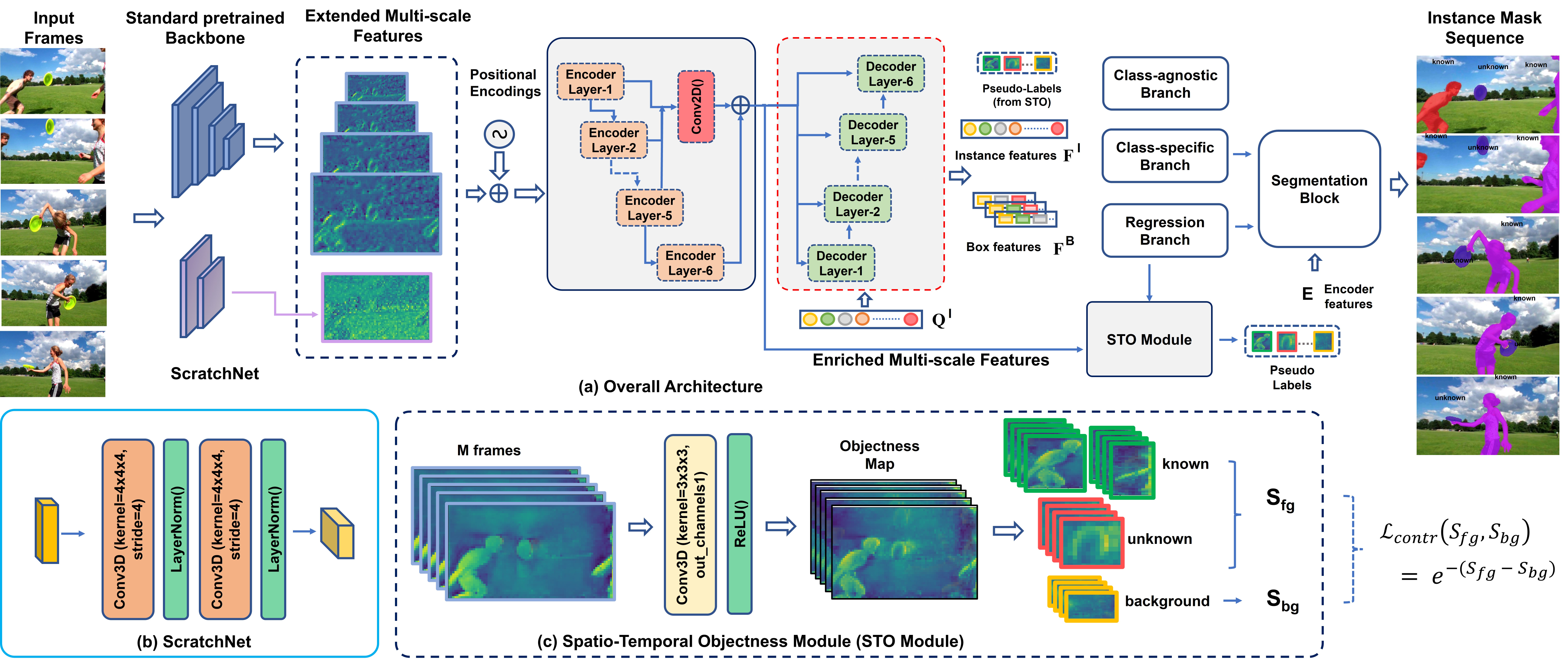}
    \caption{\textbf{(a) Overall architecture of the proposed OW-VISFormer framework.} It comprises a standard pre-trained backbone, a light-weight Scratch-Net, an encoder-decoder followed by class-specific,  class-agnostic and regression branches along with a spatio-temporal objectness (STO) module. Different from the pre-trained backbone, (b) the Scratch-Net is trained from scratch with random weight initialization during the OW-VIS training. Its light-weight architecture comprises two 3D convolution and normalization layers. The resulting feature maps are then integrated with the standard backbone features to construct extended multi-scale features which are then input to the encoder that outputs enriched features. Within the decoder, the instance queries cross-attend to the enriched features. The instance and box features output by the decoder are input to the class-specific (multi-class), class-agnostic and regression branches.  To effectively learn the class-agnostic and class-specific branches with the unknown instances in the OW-VIS setting, we introduce a (c) spatio-temporal objectness (STO) module trained with a contrastive loss ($\mathcal{L}_{contr}$) for generating instance-level pseudo-labels. Consequently, the instance features from the decoder along with the encoder features are used within the segmentation block for the video instance mask prediction of known and unknown classes in the proposed OW-VISFormer framework.
    \vspace{-0.3cm}
    } 
    \label{fig:overall_architecture}
\end{figure*}

\section{Proposed OW-VIS Framework}
\label{sec:approach}

\noindent\textbf{Overall Architecture:} Fig.~\ref{fig:overall_architecture}(a) shows the overall OW-VISFormer framework with a standard pre-trained backbone.
To circumvent using a fully-supervised ImageNet pre-trained network, we use the popular self-supervised ResNet50 DINO~\cite{dino} ImageNet-1K backbone.
For accurate pixel-level object delineation from the background, we introduce a feature enrichment mechanism that utilizes a novel light-weight ScratchNet Fig.~\ref{fig:overall_architecture}(b) which is trained from scratch through random weight initialization. Our ScratchNet, comprising two 3D convolution and normalization layers, takes the same input frames as the standard pre-trained backbone stream.  The resulting shallow features from ScratchNet are integrated with high-level semantic features from the standard pre-trained backbone to produce \textit{extended multi-scale features}. These extended features are then input to the encoder to obtain \textit{enriched multi-scale features}. The decoder within our OW-VISFormer framework aggregates the enriched features from all encoder layers, which are then cross-attended with the instance queries $\bm{Q}^I$. The decoder then outputs the instance and box features ($\bm{F}^I$ and $\bm{F}^B$), which are input to the three branches: class-specific, class-agnostic and regression. As discussed earlier, in the OW-VIS setting both the class-specific as well as the class-agnostic branches are required to be learned with the unknown instances. To this end, we propose a spatio-temporal objectness (STO) module, shown in Fig.~\ref{fig:overall_architecture}(c), comprising a 3D convolution layer for spatio-temporally aggregating the objectness information of video instances across multiple frames $M$. The STO module is trained with a contrastive loss ($\mathcal{L}_{contr}$) and enables improved foreground-background separability resulting in a better unknown instance mask prediction. Finally, the instance features from the decoder and enriched  multi-scale features from the encoder are utilized within the segmentation block to produce video instance mask predictions for both the known and unknown classes. Next, we describe our feature enrichment mechanism.

\subsection{Feature Enrichment Mechanism} 
\label{sec:scratchnet}
\noindent\textbf{Light-weight ScratchNet:} 
the OW-VIS setting requires accurate \textit{pixel-level} (unknown) object delineation from the background as well as distinguishing category-specific known classes. Therefore, shallow features capturing distinct edge and boundary information are desired to complement the high-level semantic features. 
A straightforward strategy to integrate the shallow feature information is to re-use the initial layer features from the standard pre-trained backbone. However, we empirically observe this to achieve inferior performance compared to the features generated using the proposed ScratchNet. We conjecture that the low-/mid-level shallow features from the initial layers of the standard pre-trained backbone are better adapted for the task of image classification, which prefers translation invariance. In contrast, the proposed light-weight ScratchNet produces dedicated shallow features that aid in accurate video instance mask prediction for both unknown as well as known categories.
The light-weight ScratchNet is trained from \textit{scratch} through random weights initialization. 
It comprises two layers of 3D convolution, 
each followed by a layer normalization, as shown in Fig.~\ref{fig:overall_architecture}(b). Both convolutional layers perform a non-overlapping  convolution operation on their inputs with a kernel size of $4\times4\times4$ and a stride of 4.  
ScratchNet produces complementary shallow features, which are then integrated as an additional feature scale along with multi-scale pre-trained backbone features, resulting in extended multi-scale features. \\
\noindent\textbf{Enriched Multi-scale Encoder Features:}
The extended multi-scale features are combined with positional encodings and input to a deformable encoder~\cite{Zhu_DeformableDETR_ICLR_2021} consisting of six layers of multi-scale deformable attention. The standard deformable encoder in our OW-VIS baseline (Sec.~\ref{sec:baseline}) outputs the attended multi-scale features from its final layer alone, which are tailored for known category mask prediction. This can likely lead to deterioration in some of the relevant features for accurate mask prediction of unknown instances in our OW-VIS setting. To alleviate this issue, we combine features from all encoder layers in order to learn enriched features for predicting accurate video masks for both known and unknown instances. To this end, multi-scale features from all but the final encoder layer are scale-wise fused through a convolution operation and added with the features of the final encoder layer, resulting in enriched multi-scale features. Such a feature fusion enables improving unknown video instance mask prediction while preserving the known category  predictions. Consequently, the encoder outputs enriched multi-scale features that are better suited for the OW-VIS task. These enriched multi-scale features are then input to the decoder as well as our spatio-temporal objectness (STO) module described next. 

\begin{figure*}[t!]
    \begin{center}
       \includegraphics[width=\linewidth]{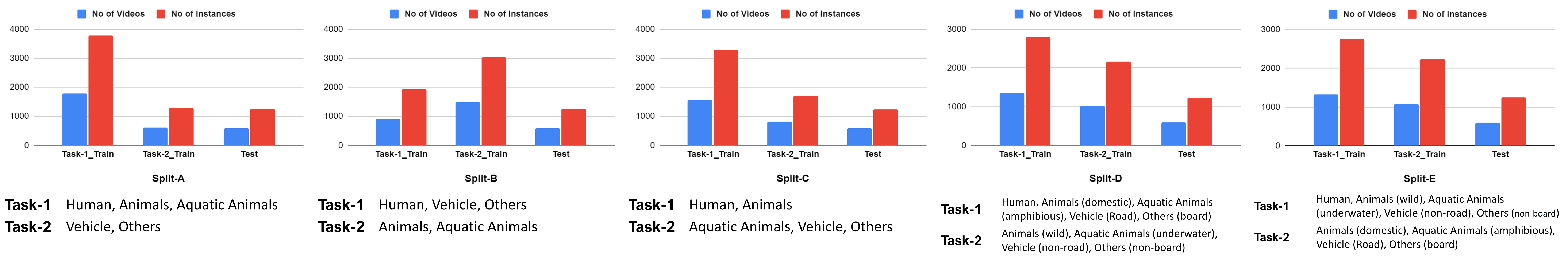} \vspace{-0.9cm}
    \end{center}
    \caption{\textbf{The proposed task composition in our OW-VIS evaluation setting based on the Youtube-VIS dataset.} Here, for each task in the corresponding split we show the number of videos and instances (objects). We construct five splits (A-E) each having two tasks (Task-1 and Task-2). For \textbf{Task-1}, the super-categories in Split A, B, C, D and E are: \{\textit{Human}, \textit{Animals}, \textit{Aquatic Animals}\}, \{\textit{Human}, \textit{Vehicle} and \textit{Others}\}, \{\textit{Human}, \textit{Animals}\}, \{\textit{Human}, \textit{Animals (domestic)}, \textit{Aquatic Animals (amphibious)}, \textit{Vehicle (road)} and \textit{Others (board)}\} and \{\textit{Human}, \textit{Animals (wild)}, \textit{Aquatic Animals (underwater)}, \textit{Vehicle (non-road)} and \textit{Others (non-board)}\}. Similarly, for \textbf{Task-2}, the super-classes in Split A, B, C, D and E are: \{\textit{Vehicle} and \textit{Others}\}, \{\textit{Animals}, \textit{Aquatic Animals}\}, \{\textit{Aquatic Animals}, \textit{Vehicle} and \textit{Others}\}, \{\textit{Human}, \textit{Animals (wild)}, \textit{Aquatic Animals (underwater)}, \textit{Vehicle (non-road)} and \textit{Others (non-board)}\}, \{\textit{Human}, \textit{Animals (domestic)}, \textit{Aquatic Animals (amphibious)}, \textit{Vehicle (road)} and \textit{Others (board)}\}. 
   \vspace{-0.2cm}
    } 
    \label{fig:split_stats}
\end{figure*}

\subsection{Spatio-Temporal Objectness Module}
\label{sec:objectness_module}

In the OW-VIS setting, both category-specific as well as class-agnostic branches require learning with the unknown instances. To this end, we introduce a spatio-temporal objectness (STO) module (Fig.~\ref{fig:overall_architecture} (c)) that generates instance-level pseudo-labels and is trained using a contrastive loss. The STO module $G_{STO}(\cdot)$ consists of a 3D convolutional layer with output channels equal to one. It takes the $d$-dimensional enriched encoder features $\bm{E}_k$ corresponding to the $M$ frames at spatial scale $k=\nicefrac{1}{16}$ as input and outputs an objectness map $O_{map} \in \mathcal{R}^{M \times\nicefrac{H}{16} \times \nicefrac{W}{16}}$. \\
\noindent\textbf{Pseudo-labeling Unknown Instances:} Given the $q$ instance features $\bm{F}^I$ output by the decoder, we employ the Hungarian matching loss~\cite{hungarian_method} that identifies the best matching queries for the $K$ known instances in the input video. For each of the remaining $q-K$ instance predictions, their objectness scores $s_i$ (where $i\in\{K+1,\cdots, q\}$) are computed by spatio-temporally aggregating the activation strengths of the objectness maps $O_{map} = G_{STO}(\bm{E}_k)$ within the corresponding predicted box regions across $m = [1,\cdots,M]$ frames, given by
\begin{equation} 
\label{eq:obj_score}
    s_i = \sum_{m=1}^{M} {\frac{1}{h_i^m\cdot w_i^m}} \sum_{x_i^m - {0.5w_i^m}}^{x_i^m + {0.5w_i^m}} \sum_{y_i^m - {0.5h_i^m}}^{y_i^m + {0.5h_i^m}} G_{STO}(\bm{E}_k),
\end{equation}

where $\bm{b}_i^m = [x_i^m, y_i^m, w_i^m, h_i^m]$ denotes the box proposal predicted in the regression branch for the $i^{th}$ instance in $m^{th}$ frame. Here, ($x_i^m, y_i^m$), $w_i^m$ and $h_i^m$ denote the center, width and height, respectively.
The resulting $q-K$ scores ($s_i$) are sorted in decreasing order and the top-$p_u$ instances are employed as pseudo-unknown during training. Furthermore, the remaining $q-(K+p_u)$ instances are considered as background instances.

Given that the selection of pseudo-unknowns depends on the activations in the objectness map $O_{map}$, 
we introduce a contrastive loss that aims to better separate the class-agnostic foreground regions from the background in the objectness map. The foreground score $S_{fg}$ is computed by aggregating the objectness scores $s_i$ of the foreground video instances, \ie, $i \in [1, \cdots, K + p_u]$ (both known and pseudo-unknown). Similarly, the background score $S_{bg}$ is obtained by aggregating the objectness scores $s_i$ of the background video instances, \ie, $i \in [K + p_u+1, \cdots, q]$. The constrastive loss is then given by
\begin{align} 
\label{eq:contrastive_loss}
    & \mathcal{L}_{contr}\bigl(S_{fg},S_{bg}\bigr) = e^{-( S_{fg} -  S_{bg})}, \\
    & \text{where} \quad S_{fg} = \sum_{i=1}^{K+p_u} s_i \quad \text{and} \quad S_{bg} = \sum_{i=K+p_u+1}^{q} s_i.
\end{align}
The resulting pseudo-unknown and background instances along with the ground-truth known instances are employed for training the class-specific and class-agnostic branches. Furthermore, as in~\cite{seqformer}, the instance and box features ($\bm{F}^I$ and $\bm{F}^B$) output by the decoder, along with the enriched multi-scale features $\bm{E}$ are utilized as input to the segmentation block for predicting the video masks of both known and unknown instances.

\subsection{Training and Inference}
\label{sec:training}
\noindent\textbf{Training:} Our proposed OW-VISFormer framework is trained with the loss formulation given by
\begin{equation} 
    \mathcal{L} = \mathcal{L}_{c} + \mathcal{L}_{r} + \alpha\mathcal{L}_{f} + \mathcal{L}_{contr},
\end{equation}
where $\mathcal{L}_{c}$, $\mathcal{L}_{f}$, $\mathcal{L}_{r}$ and $\mathcal{L}_{contr}$, respectively denote the classification (class-specific branch), foreground objectness (class-agnostic branch), regression (box and mask) and constrastive (STO module) loss terms. While $\mathcal{L}_{c}$ and $\mathcal{L}_{f}$ are computed using the focal loss~\cite{FocalLoss}, $\mathcal{L}_{r}$ is the standard $l_{1}$ loss. 
Furthermore, a balanced set of exemplars is utilized to finetune the model after the incremental step in each task for alleviating catastrophic forgetting, as in~\cite{ow-detr,joseph2021towards}.\\

\noindent\textbf{Inference:} At test time, the class-specific branch predictions of the $C$ known classes are utilized and top-$k$ instances are selected as known instances. Furthermore, among the remaining $q-k$ instances, top-$k$ with high unknown class probability are selected as unknown instances. Finally, the box and instance features from the decoder for the corresponding known and unknown instances predicted, along with the enriched multi-scale encoder features are input to the segmentation block for predicting the video masks.

\begin{table*}[t!]
\caption{\textbf{Comparison between the baseline and our OW-VISFormer on the five splits (A-E) introduced for OW-VIS setting.} For the Task-1 evaluation, the results are reported in terms of overall AP and recall (AR-1) for both `Known' classes and the `Unknown' category. For the Task-2, which involves the incremental learning step, we report the OW-VIS setting results for `Previously Known', `Current Known' along with 'Both'. Note that the `Unknown' class performance is not reported for Task-2 since all 40 classes are `Known'. Our proposed OW-VISFormer  achieves consistent gains over the baseline on all splits across all tasks for both 'Known' and `Unknown' classes.
}
\label{table:owvis}
\centering
\setlength{\tabcolsep}{14pt}
\adjustbox{width=0.96\textwidth}{
    \begin{tabular}{cc|cccc|cccccc}
    \toprule
    &  & \multicolumn{4}{c|}{\textbf{Task-1}} & \multicolumn{6}{c}{\textbf{Task-2}} \\
    \cmidrule{3-12}
    &  & \multicolumn{2}{c}{\cellcolor{gray}\textit{Known}} & \multicolumn{2}{c|}{\cellcolor{gray}\textit{Unknown}} & \multicolumn{2}{c}{\cellcolor{gray}\textit{Previously Known}} & \multicolumn{2}{c}{\cellcolor{gray}\textit{Current Known}} & \multicolumn{2}{c}{\cellcolor{gray}\textit{Both}} \\
    &  & \cellcolor{gray}AP & \cellcolor{gray}AR-1 & \cellcolor{gray}AP & \cellcolor{gray}AR-1 & \cellcolor{gray}AP & \cellcolor{gray}AR-1 & \cellcolor{gray}AP & \cellcolor{gray}AR-1 & \cellcolor{gray}AP & \cellcolor{gray}AR-1 \\ \midrule
    \multirow{2}{*}{ \textbf{Split A }} & \texttt{Baseline} & 35.3 & 34.8 & 6.7 & 9.5 & 30.6 & 31.2 & 30.3 & 30.7 & 30.4 & 31.0 \\
    & \cellcolor{orange!6} \texttt{Ours} & \cellcolor{orange!6} 36.7 & \cellcolor{orange!6} 37.3 & \cellcolor{orange!6} 10.0 & \cellcolor{orange!6} 11.9 & \cellcolor{orange!6} 32.5 & \cellcolor{orange!6} 33.9 & \cellcolor{orange!6} 34.1 & \cellcolor{orange!6} 34.9 & \cellcolor{orange!6} 33.3 & \cellcolor{orange!6} 34.4 \\ \midrule
    \multirow{2}{*}{ \textbf{Split B } } & \texttt{Baseline} & 31.0 & 31.4 & 2.7 & 5.1 & 28.9 & 30.5 & 31.7 & 32.6 & 30.3 & 31.6 \\
    & \cellcolor{orange!6} \texttt{Ours} & \cellcolor{orange!6} 32.2 & \cellcolor{orange!6} 33.1 & \cellcolor{orange!6} 6.5 & \cellcolor{orange!6} 8.7 & \cellcolor{orange!6} 30.4 & \cellcolor{orange!6} 32.2 & \cellcolor{orange!6} 35.1 & \cellcolor{orange!6} 35.7 & \cellcolor{orange!6} 32.7 & \cellcolor{orange!6} 33.9 \\ \midrule
    \multirow{2}{*}{ \textbf{Split C } } & \texttt{Baseline} & 33.9 & 33.3 & 4 & 7.2 & 28.7 & 32.6 & 32.1 & 32.6 & 30.4 & 28.7 \\
    & \cellcolor{orange!6} \texttt{Ours} & \cellcolor{orange!6} 36.4 & \cellcolor{orange!6} 35.2 & \cellcolor{orange!6} 7.1 & \cellcolor{orange!6} 9.6 & \cellcolor{orange!6} 30.6 & \cellcolor{orange!6} 33.2 & \cellcolor{orange!6} 35.0 & \cellcolor{orange!6} 35.1 & \cellcolor{orange!6} 32.8 & \cellcolor{orange!6} 34.1 \\ \midrule
    \multirow{2}{*}{ \textbf{Split D } } & \texttt{Baseline} & 31.4 & 34.5 & 3.3 & 6.4 & 29.7 & 30.9 & 30.2 & 32.2 & 30.0 & 31.6 \\
    & \cellcolor{orange!6} \texttt{Ours} & \cellcolor{orange!6} 33.6 & \cellcolor{orange!6} 35.0 & \cellcolor{orange!6} 6.9 & \cellcolor{orange!6} 9.7 & \cellcolor{orange!6} 31.7 & \cellcolor{orange!6} 32.2 & \cellcolor{orange!6} 33.5 & \cellcolor{orange!6} 34.2 & \cellcolor{orange!6} 32.6 & \cellcolor{orange!6} 33.2 \\ \midrule
    \multirow{2}{*}{ \textbf{Split E } } & \texttt{Baseline} & 32.0 & 35.1 & 3.5 & 6.5 & 29.7 & 31.2 & 30.4 & 31.7 & 30.1 & 29.7 \\
    & \cellcolor{orange!6} \texttt{Ours} & \cellcolor{orange!6} 35.1 & \cellcolor{orange!6} 36.3 & \cellcolor{orange!6} 5.6 & \cellcolor{orange!6} 8.9 & \cellcolor{orange!6} 31.3 & \cellcolor{orange!6} 32.1 & \cellcolor{orange!6} 33.9 & \cellcolor{orange!6} 36.0 & \cellcolor{orange!6} 32.6 & \cellcolor{orange!6} 34.0 \\ \bottomrule
    \end{tabular}
    \vspace{-0.5cm}
}
\end{table*}

\section{Experiments}
\subsection{Experiment 1: OW-VIS Setting}
\noindent \textbf{Datasets:} We adapt the popular Youtube-VIS \cite{YouTube-VIS-2021}  dataset to construct open-world VIS (OW-VIS) data splits. For each split, we group the $40$ categories into two mutually exclusive sets, thereby constructing tasks with non-overlapping classes $\{\mathcal{T}_1, \mathcal{T}_2\}$ such that, the $\mathcal{T}_{2}$ categories are not known during task $\mathcal{T}_{1}$.
Then, during the learning of task $\mathcal{T}_{t}$, all the category labels belonging in $\{\mathcal{T}_{\alpha} : \alpha \leq t\}$ are considered as \textit{known}. Similarly, labels belonging to $\{\mathcal{T}_{\alpha} : \alpha > t\}$ are considered as \textit{unknown} during evaluation. To construct a test set at time $t$, we consider 20\% of videos in tasks $\{\mathcal{T}_{\alpha} : \alpha \leq t\}$ for \textit{known} evaluation and 20\% of videos in tasks $\{\mathcal{T}_{\alpha} : \alpha > t\}$ for \textit{unknown} evaluation. By grouping super-categories of all 40 Youtube-VIS categories in different ways, we created $5$ such splits as shown in Fig.~\ref{fig:split_stats}. Additional details are provided in the supplementary.\\
\noindent \textbf{Evaluation Metric:}
We adapt the standard VIS evaluation metrics for evaluating the OW-VIS setting. For the known classes as well as the unknown class, the standard overall average precision (AP) and average recall (AR) are used.

\noindent \textbf{Implementation Details:} We employ the self-supervised DINO~\cite{dino} ResNet50 Imagenet-1k pre-trained backbone to extract multi-scale features using the $conv_{3}$, $conv_{4}$ and $conv_{5}$ stages. The resulting multi-scale features are mapped to the same feature dimension of $256$ through convolution, as in~\cite{seqformer,Zhu_DeformableDETR_ICLR_2021}. The two convolution layers in ScratchNet employ kernel size $4\times 4$ and stride $4$ along with the output channels set to $256$. The  encoder and decoder are six layers each with latent dimension being $256$. The number of instance queries $q$ is set to $300$. Our OW-VISFormer is implemented in PyTorch-1.8~\cite{pytorch} and learned using the Adam optimizer with learning rate set to $10^{-4}$. The model is trained on Task-1 data for $18$ epochs. For Task-2, the learned model from Task-1 is incrementally trained via the memory replay. Specifically, it is trained on current known classes of Task-2 for $12$ epochs and then finetuned for $2$ epochs by including previous known class exemplars. Our code, models and data splits will be made public.

\noindent \textbf{Quantitative Comparison:}  We compare the performance of the baseline (\ref{sec:baseline}) and our OW-VISFormer (\ref{sec:approach}) on the OW-VIS splits (see Fig.~\ref{fig:split_stats}). Tab.~\ref{table:owvis} shows the comparison on all five splits and the corresponding two tasks in each split. We report the performance in terms of AP and recall for both known and unknown. For the Task-1 in split A, the baseline achieves a known class AP of 35.3\% and AR-1 score of 34.8\%. Our OW-VISFormer achieves consistent improvement in performance in terms of both AP and AR-1 by achieving 36.7\% and 37.3\%, respectively.  Notably, OW-VISFormer obtains a considerable gain in performance in the case of unknown class, in terms of both AP and AR-1, owing to the proposed feature enrichment mechanism and the STO module. When the unknown class labels are progressively labeled in Task-2, we observe OW-VISFormer to better maintain both objectives of (i) spatio-temporally segmenting the new known categories and (ii) not forgetting the previously known classes. Moreover, we observe a consistent improvement in performance from OW-VISFormer over the baseline for other splits. 

\begin{figure*}[t!]
    \centering
      \includegraphics[width=0.98\linewidth]{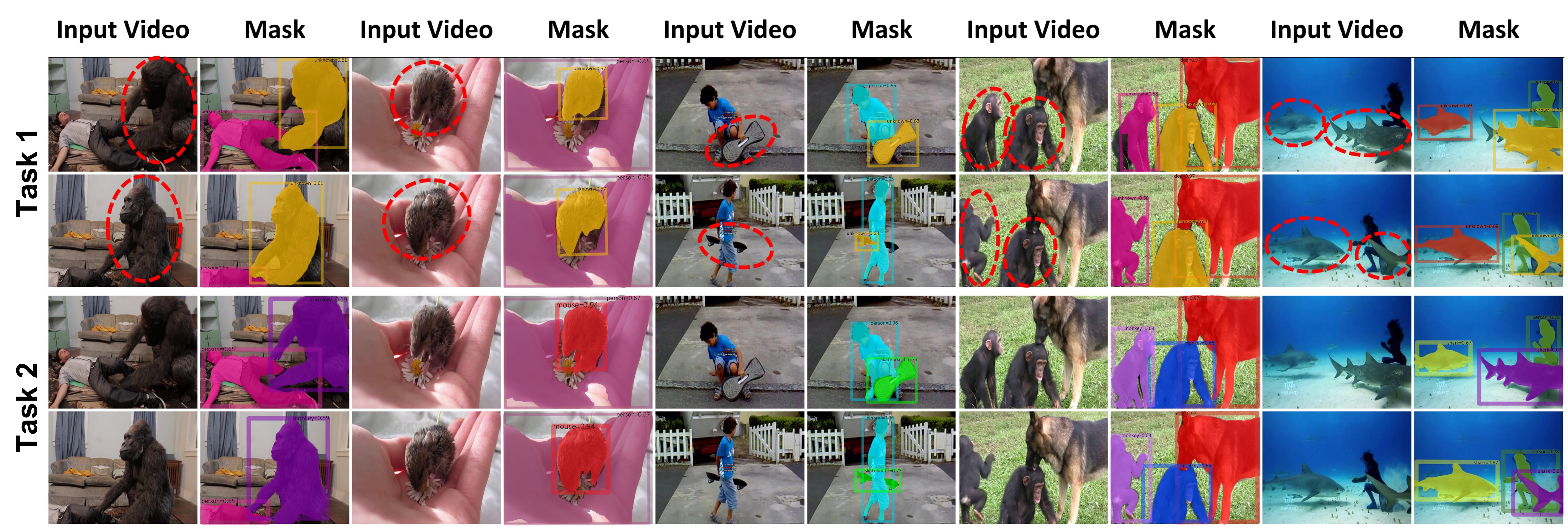}
     \vspace{-0.1cm}
    \caption{\textbf{Qualitative OW-VIS results on example video frames from the the test sets of different splits}. Here, for each example video, we show the segmentation masks obtained from our 
    OW-VISFormer when trained only on Task-1 categories (row 1 and 2). The instance mask predictions for the same video frames are shown after \textit{incrementally} learning with Task-2 categories (row 3 and 4). From left to right: the unknown objects are encircled by red dashed lines in the input video frames during the Task-1 evaluation. The unknown objects are accurately segmented first as `unknown' in Task-1 and then later correctly classified and spatio-temporally segmented into their respective `known' classes during Task-2 evaluation. Best viewed zoomed in. Additional results are in the supplementary.
      \vspace{-0.3cm}
    } 
    \label{fig:qual_results}
\end{figure*}
\begin{table}[t!]
\centering
  \caption{\textbf{State-of-the-art comparison on YouTube-VIS 2019 val set}. All results are reported using the same ResNet-50 backbone. Our approach outperforms the recent SeqFormer \cite{seqformer} with an absolute gain of 1.6\% in terms of overall AP. Best results are in bold.
  }
  \label{table:supervised_results}

  \scalebox{0.81}{
  \begin{tabular}{l|ccccc}
  \toprule
  \rowcolor{gray}
  \textbf{Method} 
  & \textbf{AP} 
  & \textbf{AP}$_{\mathtt{50}}$ 
  & \textbf{AP}$_{\mathtt{75}}$ 
  & \textbf{AR}$_{\mathtt{1}}$ 
  & \textbf{AR}$_{\mathtt{10}}$ 
  \\
  \toprule
  MaskTrack R-CNN\cite{YouTube-VIS-2019} & $30.3$ & $51.1$ & $32.6$ & $31.0$ & $35.5$ \\
  SipMask-VIS\cite{SipMask} & $32.5$ & $53.0$ & $33.3$ & $33.5$ & $38.9$ \\
  VisTR\cite{vistr} & $36.2$ & $59.8$ & $36.9$ & $37.2$ & $42.4$ \\
  CrossVIS\cite{CROSS-VIS} & $36.6$ & $57.3$ & $39.7$ & $36.0$ & $42.0$ \\
  IFC\cite{ifc} & $42.8$ & $65.8$ & $46.8$ & $43.8$ & $51.2$ \\
  DeVIS\cite{devis} & $44.4$ & $66.7$ & $48.6$ & $42.4$ & $51.6$ \\
  SeqFormer\cite{seqformer} & $45.1$ & $66.9$ & $50.5$ & $45.6$ & $54.1$ \\
  \rowcolor{orange!6}
  \textbf{Our Approach} & \textbf{46.7} & \textbf{69.1} & \textbf{51.7} & \textbf{46.1} & \textbf{54.9} \\
  \bottomrule
  \end{tabular}
 }
 \vspace{-0.7cm}
 
\end{table}
\noindent\textbf{Ablation Study:} We also conduct an experiment by progressively integrating our contributions into the baseline on one of the difficult splits (split B). For the known classes, the baseline achieves a AP score of 31.0\%. The performance on the known classes is improved by the introduction of our  proposed feature enrichment mechanism (Sec.~\ref{sec:scratchnet}) with an AP score of 31.5\%. 
Furthermore, the integration of our STO module (Sec.~\ref{sec:objectness_module}) that generates instance-level pseudo-labels improves the AP score to 32.2\%, leading to an absolute overall gain of 1.2\% over the baseline. For the unknown class, the baseline achieves AP and AR-1 scores of 2.7\% and 5.1\%. Integrating the proposed feature enrichment mechanism significantly improves the performance to 4.5\% and 6.1\%, in terms of AP and AR-1. The introduction of the STO module leads to a consistent improvement in both AP and AR-1, achieving absolute final gains of 3.8\% and 3.6\% over the baseline. We further perform an  experiment to validate the impact of the dedicated shallow features generated from the proposed ScratchNet. To this end, we use the shallow features $conv_2$ from the standard pre-trained backbone instead in our OW-VISFormer. The reduces the performance in terms of both known and unknown AP from 32.2\% and 6.5\% to 31.2\% and 3.8\%. This inferior performance likely suggests that the shallow features from standard pre-trained backbone are more suited for image classification task. In contrast, the shallow features obtained through our ScratchNet are trained from scratch on OW-VIS data and therefore dedicated to aid in accurate video mask prediction for both unknown as well as known classes. 

\noindent \textbf{Qualitative Analysis:} Fig.~\ref{fig:qual_results} shows the qualitative results from our OW-VISFormer on example test video frames of different OW-VIS splits. The first two rows shows the segmentation results obtained on the corresponding video frames in the Task-1 evaluation (`known' and `unknown' objects). The last two rows show the results from Task-2 evaluation which involves incrementally training with Task-2 categories. In the Task-1, OW-VISFormer is able to first accurately segment the different `known' and `unknown' instances. 
Then, in the Task-2 evaluation, OW-VISFormer successfully identifies the correct categories for the same unknown instances that were introduced in the Task-2 learning while still accurately predicting the instance masks for the previously known categories from Task-1.

\begin{table*}[t!]
\scriptsize
\centering
 \caption{\textbf{State-of-the-art comparison for the open-world object detection (OWOD) problem on MS COCO split of \cite{ow-detr}.} The comparison is presented in terms of unknown category recall (U-Recall) and the known class mAP for the Task-1. For the remaining tasks (2-4) involving incremental learning steps, we report the mAP scores for `Previously Known', `Current Known' along with `Both'. Furthermore, the U-Recall is only reported for tasks 1-3 since all classes are known in the final task 4. 
 When using the same backbone, our approach outperforms the recent OW-DETR on all tasks for both the `Known` classes and the `Unknown' class. Best results are in bold.
 }
 \label{table:detection_results}
 
 \adjustbox{width=1\textwidth}{
    \begin{tabular}{l|cc|cccc|cccc|ccc}
    \toprule
    \rowcolor{gray}
    \textbf{Task IDs} & \multicolumn{2}{c|}{\textbf{Task-1}} & \multicolumn{4}{c|}{\textbf{Task-2}} & \multicolumn{4}{c|}{\textbf{Task-3}} & \multicolumn{3}{c}{\textbf{Task-4}} \\
    \toprule
    & U-Recall & mAP & U-Recall & \multicolumn{3}{c|}{mAP} & U-Recall & \multicolumn{3}{c|}{mAP} & \multicolumn{3}{c}{mAP} \\
    & \multirow{2}{*}{ } & \textit{Current} & \multirow{2}{*}{ } & \textit{Previously} & \textit{Current} & \multirow{2}{*}{ \textit{Both} } & \multirow{2}{*}{ } & \textit{Previously} & \textit{Current} & \multirow{2}{*}{ \textit{Both} } & \textit{Previously} & \textit{Current} & \multirow{2}{*}{ \textit{Both} } \\
    & & \textit{Known} & & \textit{Known} & \textit{Known} & & & \textit{Known} & \textit{Known} & & \textit{Known} & \textit{Known} & \\
    \midrule
    ORE-EBUI~\cite{joseph2021towards} & 1.5 & 61.4 & 3.9 & 56.5 & 26.1 & 40.6 & 3.6 & 38.7 & 23.7 & 33.7 & 33.6 & 26.3 & 31.8 \\
    OW-DETR~\cite{ow-detr} & 5.7 & 71.5 & 6.2 & 62.8 & 27.5 & 43.8 & 6.9 & 45.2 & 24.9 & 38.5 & 38.2 & 28.1 & 33.1\\
    \rowcolor{orange!6} \textbf{Our Approach} & \textbf{8.8} & \textbf{72.1} & \textbf{8.4} & \textbf{63.3} & \textbf{28.1} & \textbf{44.5} & \textbf{9.1} & \textbf{45.4}  & \textbf{25.5}  & \textbf{38.9}  & \textbf{38.7}  & \textbf{29.3}  & \textbf{34.0} \\   
    \bottomrule
    \end{tabular}
 }
 \vspace{-0.6cm}
\end{table*}

\subsection{Experiment 2: Fully-supervised VIS Setting}
\begin{figure}[t!]
    \centering
      \includegraphics[width=1\linewidth]{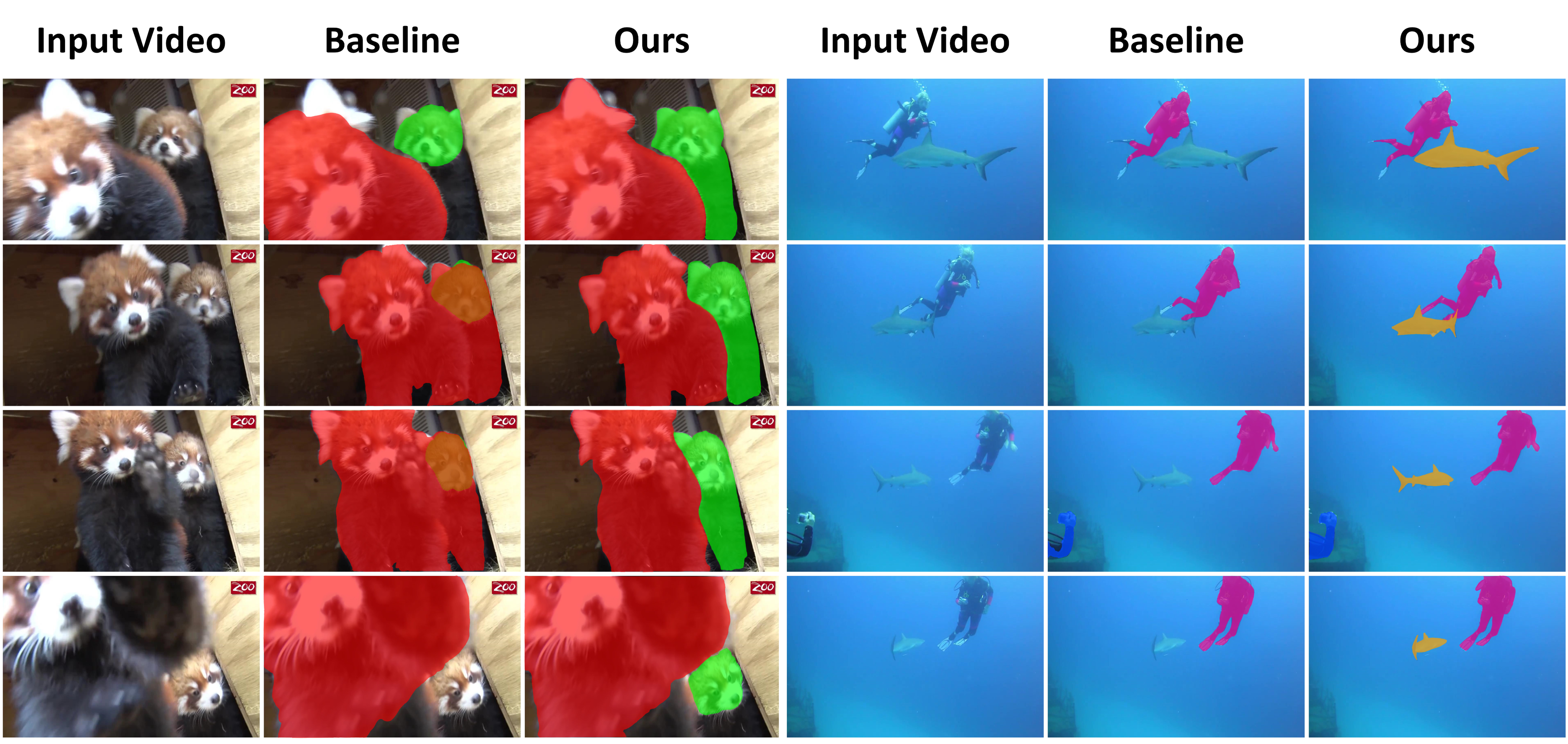}
    \caption{\textbf{Qualitative comparison between the baseline SeqFormer and our approach on the fully-supervised setting on Youtube-VIS 2019 val. set}. Our approach obtains improved instance mask segmentation, compared to the baseline. Best viewed zoomed in. Additional results are in the supplementary.
     \vspace{-0.7cm}
    } 
    \label{fig:fs_qual}
\end{figure}

We further evaluate the effectiveness of our proposed contributions (feature enrichment mechanism and STO module) in the standard full-supervised (FS) VIS problem setting. Our main intuition is that the proposed contributions can likely aid in reducing the confusion of an object region (both known and unknown) being called as a background. Hence, this can serve as an additional learning step for better object identification. 
To this end, we integrate the feature enrichment mechanism (Sec.~\ref{sec:scratchnet}) and our STO module (Sec.~\ref{sec:objectness_module}) into the recent SeqFormer \cite{seqformer}. Tab.~\ref{table:supervised_results} shows the results in the standard FS setting on the Youtube-VIS 2019 val. set. Our approach obtained by integrating the feature enrichment and the STO module within the standard SeqFormer obtains overall AP of 46.7\%, leading to an absolute gain of 1.6\% in AP over recent best~\cite{seqformer}. Fig.~\ref{fig:fs_qual} presents a qualitative comparison of our approach with the standard SeqFormer. Our approach achieves improved video instance mask predictions, compared to SeqFormer. 

\subsection{Experiment 3: Open-World Detection Setting}
Lastly, we also validate the generalizability of our feature enrichment and the STO module for open-world object detection (OWOD) in images. The OWOD problem has recently gained popularity with evaluations being performed on the challenging MS COCO dataset~\cite{MSCOCO}. To this end, we integrate our feature enrichment and STO module (replacing 3D convolutions with 2D) into the recent transformers-based OWOD framework, named OW-DETR~\cite{ow-detr}. Tab.~\ref{table:detection_results} shows the results on the challenging MS COCO split introduced in~\cite{ow-detr}. For a fair comparison, we employ the same self-supervised ResNet50 backbone as in the standard OW-DETR~\cite{ow-detr}. We report the results of ORE-EBUI~\cite{joseph2021towards} and our baseline OW-DETR from~\cite{ow-detr}.  In the case of Task-1, our approach achieves an impressive performance particularly on the `Unknown' object class by increasing the U-Recall from 5.7\% to 8.8\%, while also improving the detection performance on the `Current Known' classes. Further, the performance on both `Unknown` and `Known' are consistently improved in the subsequent tasks as well. For the final task (task-4) involving all 80 classes from MS COCO as `Known`, our approach improves the performance over the recent OW-DETR in the case of `Both' previously known and unknown classes from 33.1\% to 34.0\% mAP. 

\section{Relation to Prior Art}
Existing video instance segmentation (VIS) methods can be categorized based on the underlined detection architecture such as, two-stage \cite{YouTube-VIS-2019,VAEVIS,MaskPropagation}, single-stage \cite{PCAN,SipMask,STEmSeg,CompFeat,dudhane2021burst} and transformer-based approaches \cite{vistr,seqformer, CROSS-VIS,ifc,devis,aleissaee2022transformers}. Most two-stage VIS approaches extend a two-stage detector such as, Mask R-CNN~\cite{MaskRCNN} by integrating a tracking branch. Most single-stage VIS methods adapt the one-stage pipeline, where the final mask prediction is obtained through a linear combination of mask bases. Recently, several works explored transformers-based detection architecture \cite{Zhu_DeformableDETR_ICLR_2021,DETR} to formulate VIS as a direct end-to-end sequence prediction. These approaches predominantly solve the problem following a closed-world assumption, where the annotated instances of all (known) semantic classes to be spatio-temporally segmented are available during instance. This assumption posses issues to most existing VIS approaches when classifying a novel (unknown) object class instance.

Several recent works have explored the open-world problem setting for different tasks, such as detection \cite{joseph2021towards,ow-detr}, image instance segmentation \cite{learn2det_ow,wang2022open}, and object tracking and video segmentation \cite{open_ow_track,wang2021unidentified}. However, these works on open-world video object segmentation \cite{open_ow_track,wang2021unidentified} only look into the task of unknown identification and does not address the more challenging open-world problem setting explored in this work, which requires (i) distinguishing a set of `known' semantic classes as well as identifying an unknown object as `unknown' and then (ii) incrementally learning the class of an `unknown' as and when the corresponding semantic category labels are available. 
To the best of our knowledge, we are the first to investigate and introduce an approach for the problem of open-world VIS (OW-VIS). 

\section{Conclusions}
We proposed an approach, named OW-VISFormer, to address open-world VIS (OW-VIS). OW-VISFormer introduces a feature enrichment mechanism to produce enriched features and a spatio-temporal objectness module that generates instance-level pseudo-labels. 
Moreover, we propose OW-VIS  splits to identify unknown, segment known and unknown along with progressively segmenting new semantic classes. OW-VISFormer achieves competitive performance in three settings: OW-VIS, FS-VIS and OWOD. 

{\small
\bibliographystyle{ieee_fullname}
\bibliography{egbib}
}

\end{document}